\documentclass[sigconf]{acmart}

\AtBeginDocument{%
  }

\copyrightyear{2026}
\acmYear{2026}
\setcopyright{cc}
\setcctype{by-nc-nd}
\acmConference[SA Technical Communications '26]{SIGGRAPH Asia 2026 Technical Communications}{December 01--04, 2026}{Kuala Lumpur, Malaysia}
\acmBooktitle{SIGGRAPH Asia 2026 Technical Communications (SA Technical Communications '26), December 01--04, 2026, Kuala Lumpur, Malaysia}
\acmDOI{10.1145/3829339.3847845}
\acmISBN{979-8-4007-2841-9/2026/12}

\usepackage{subcaption}
\usepackage{multirow}
\graphicspath{{figures/}}
\usepackage{tikz}
\usetikzlibrary{arrows.meta}

\newcommand{\RESLR}{1024}   %
\newcommand{\RESHR}{4096}   %
\newcommand{\SRFAC}{4}      %
\newcommand{\NTEST}{100}

\begin{document}

\title{Domain-Tuned Diffusion Super-Resolution for 4K Facial Asset Maps}

\author{Elly Akhoundi}
\email{eakhoundi@ea.com}
\orcid{0009-0006-4989-1731}
\affiliation{%
  \institution{Electronic Arts}
  \city{Vancouver}
  \country{Canada}
}

\author{Mathieu Lamarre}
\email{mlamarre@ea.com}
\orcid{0009-0008-3965-1173}
\affiliation{%
  \institution{Electronic Arts}
  \city{Montr\'eal}
  \country{Canada}
}

\author{Thibault Goyallon}
\email{tgoyallon@ea.com}
\orcid{0000-0003-3238-3707}
\affiliation{%
  \institution{Electronic Arts}
  \city{Toronto}
  \country{Canada}
}

\author{Patrick Anderson}
\email{panderson@ea.com}
\orcid{0009-0004-4386-4789}
\affiliation{%
  \institution{Electronic Arts}
  \city{Montr\'eal}
  \country{Canada}
}

\author{\'Etienne Danvoye}
\email{edanvoye@ea.com}
\orcid{0009-0006-4837-9541}
\affiliation{%
  \institution{Electronic Arts}
  \city{Montr\'eal}
  \country{Canada}
}
\begin{abstract}
Physically based facial assets require 4K albedo and displacement maps for cinematic close-ups, yet generative pipelines commonly produce them at only 1K.
Naively applying photographic super-resolution is a poor fit: facial UV maps have domain-specific statistics, and displacement is a scalar height field whose errors directly distort geometry.
We present a domain-tuned diffusion upscaler for $\SRFAC\times$ ($\RESLR{\to}\RESHR$) facial asset super-resolution.
Starting from a bilinearly upsampled map, we adapt a frozen SDXL backbone per modality with lightweight LoRA modules, while a Tile ControlNet preserves input structure as the diffusion prior restores pore- and wrinkle-scale detail.
On held-out maps, our method produces sharper, more faithful detail than bilinear and off-the-shelf GAN and diffusion super-resolvers, reducing albedo LPIPS from $0.21$ to $0.08$ while avoiding misplaced or excessive displacement relief.
These results show that lightweight domain adaptation can produce render-ready 4K facial asset maps without a custom super-resolution architecture.

\end{abstract}



\begin{teaserfigure}
\centering
\resizebox{\textwidth}{!}{%
\begin{tabular}{@{}cccc@{}}
\footnotesize 1K albedo & \footnotesize 4K albedo (ours) & \footnotesize 1K displacement & \footnotesize 4K displacement (ours) \\
\includegraphics[width=3.35cm,trim={0 800bp 0 200bp},clip]{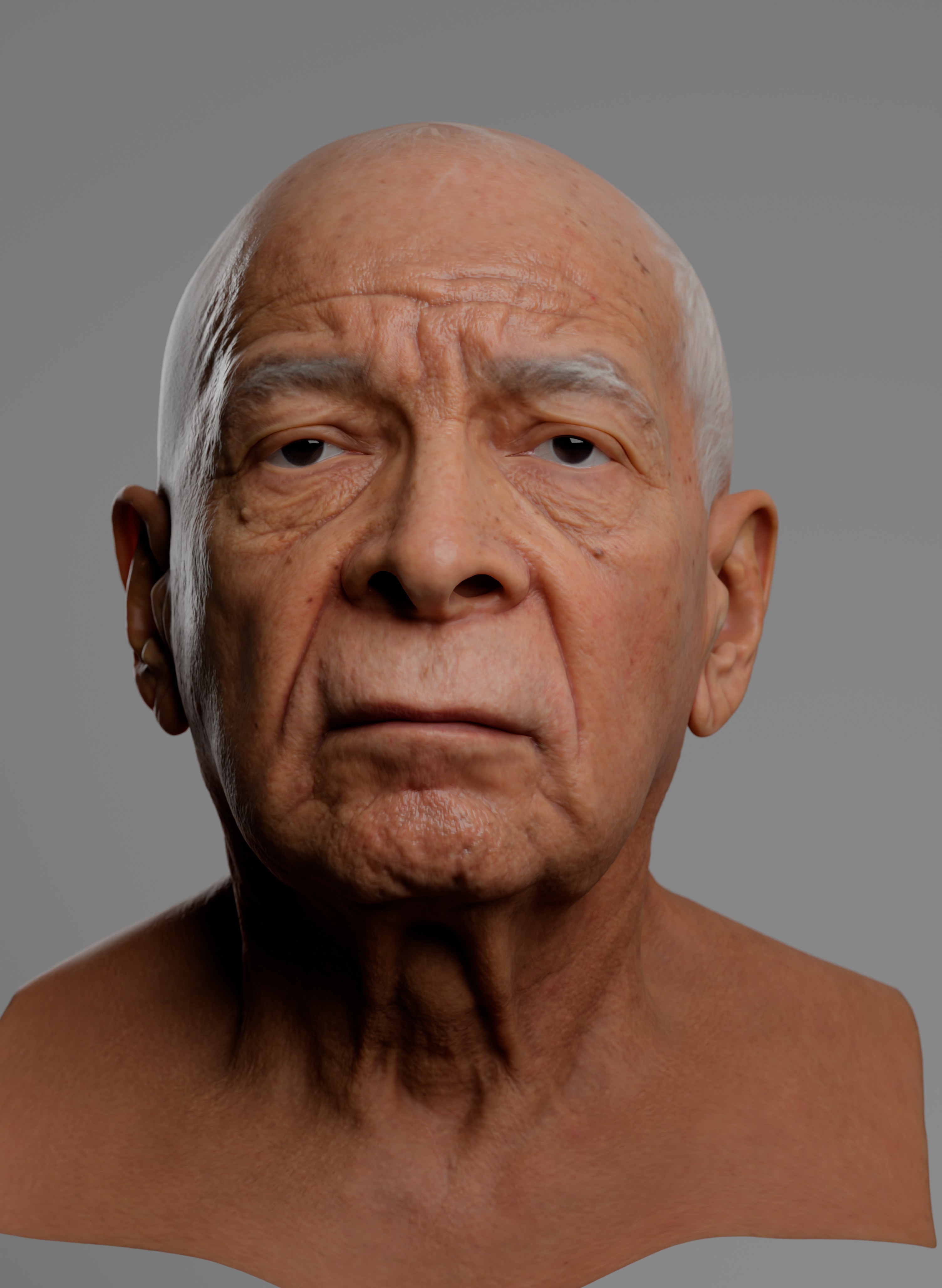} &
\includegraphics[width=3.35cm,trim={0 800bp 0 200bp},clip]{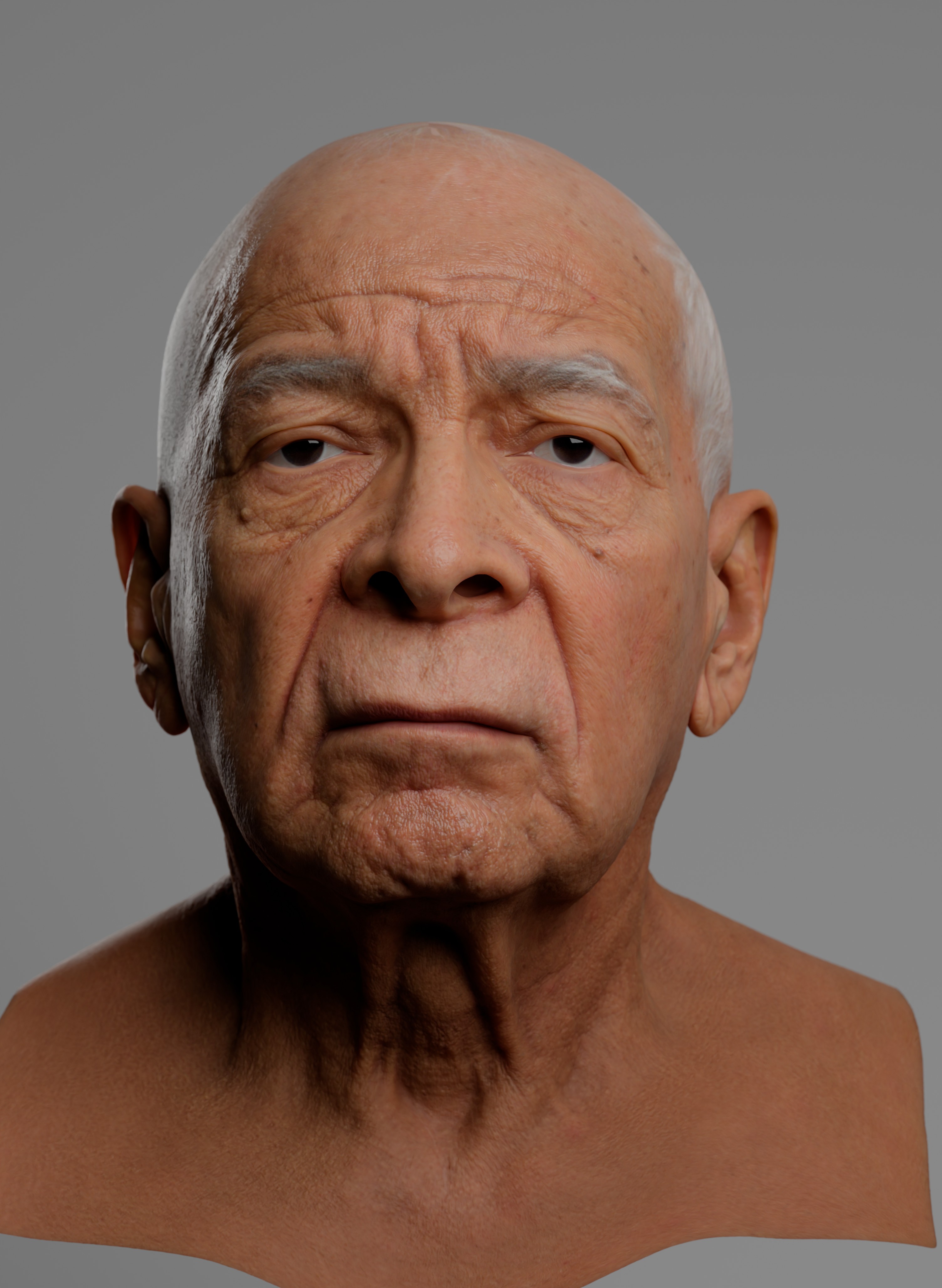} &
\includegraphics[width=3.35cm,trim={0 800bp 0 200bp},clip]{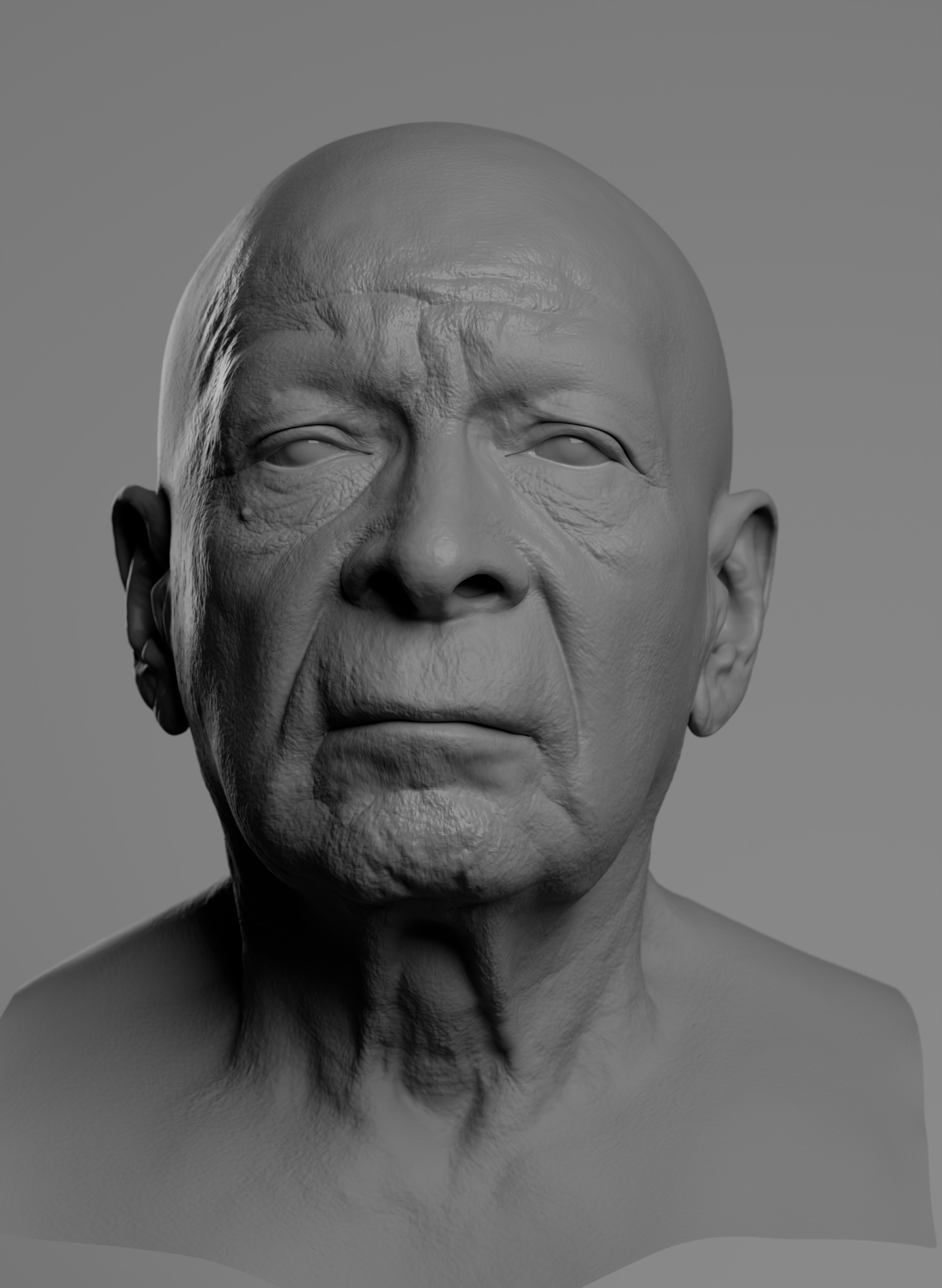} &
\includegraphics[width=3.35cm,trim={0 800bp 0 200bp},clip]{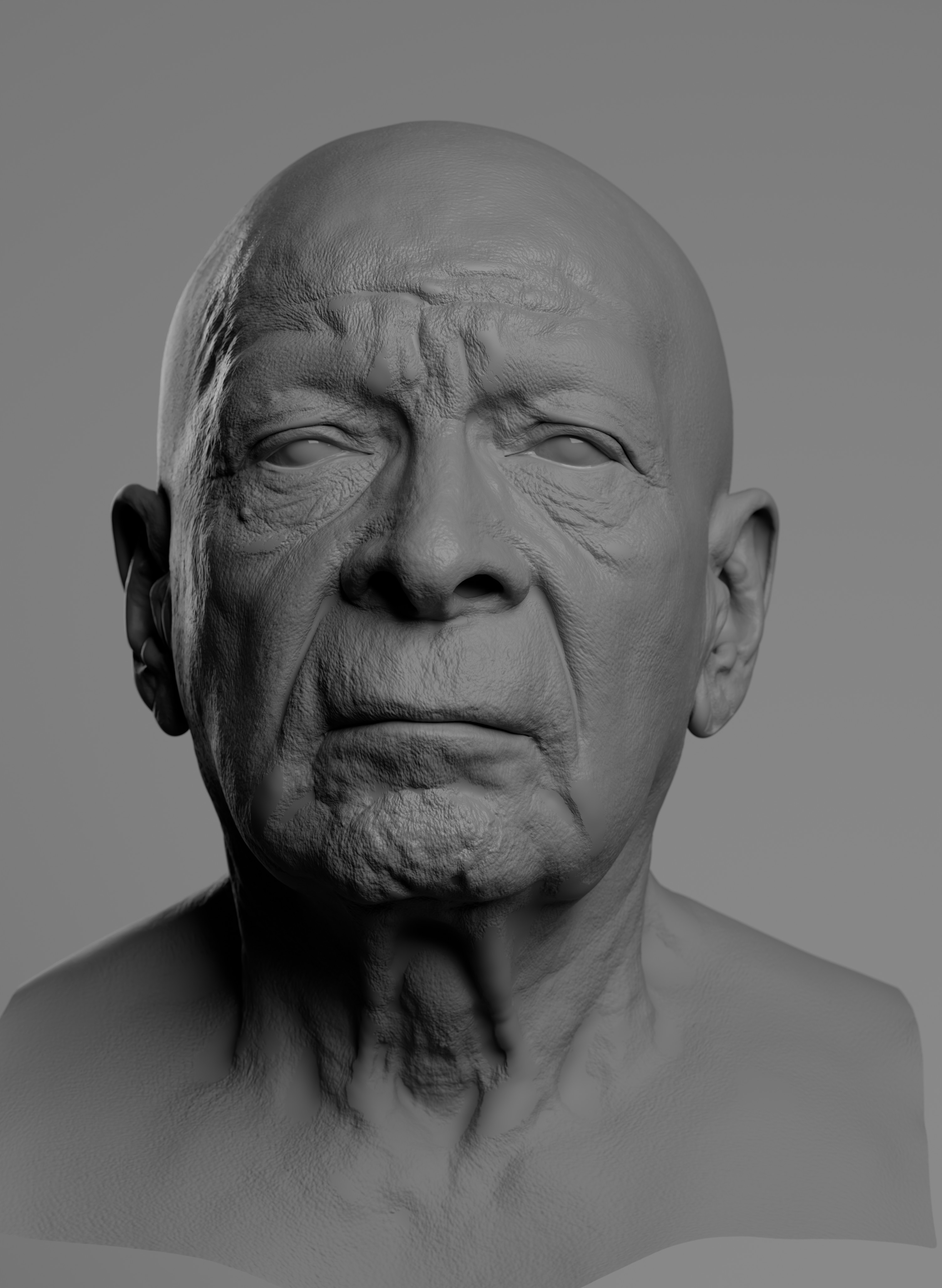} \\[4pt]
\includegraphics[width=3.35cm]{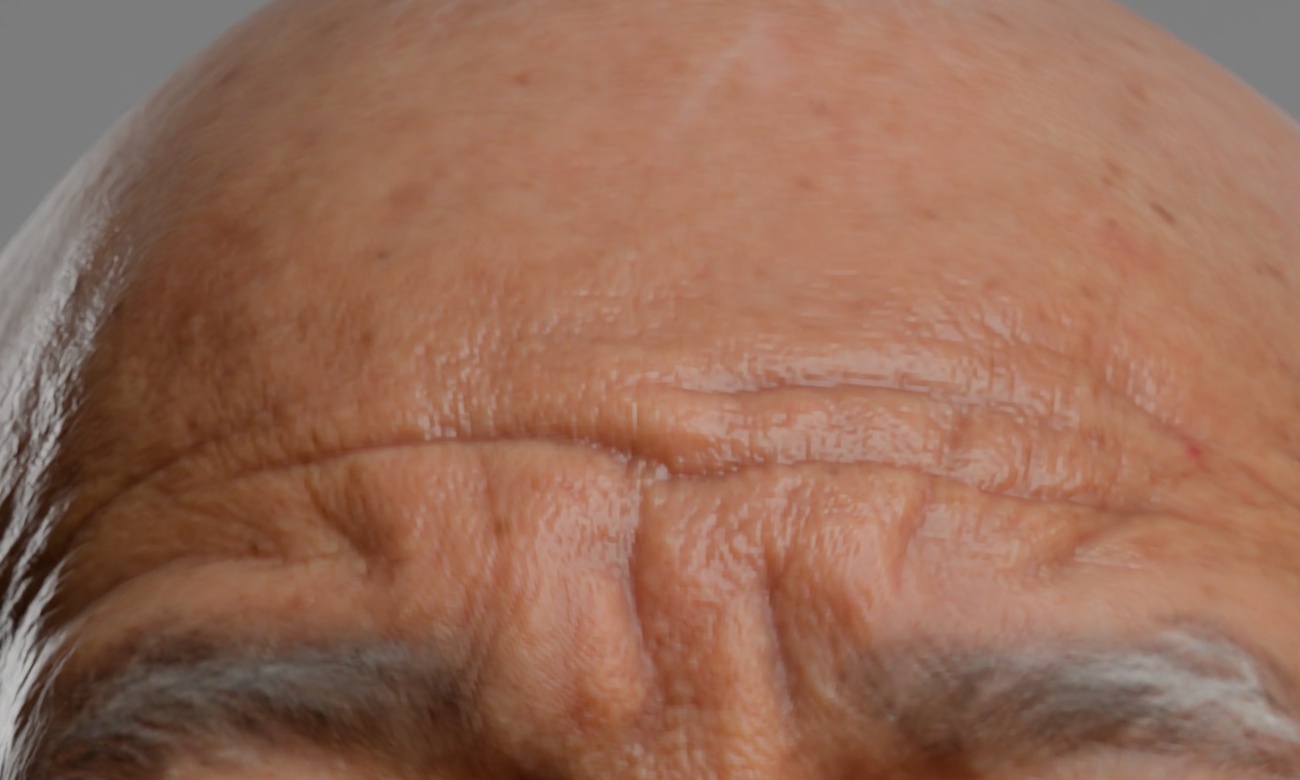} &
\includegraphics[width=3.35cm]{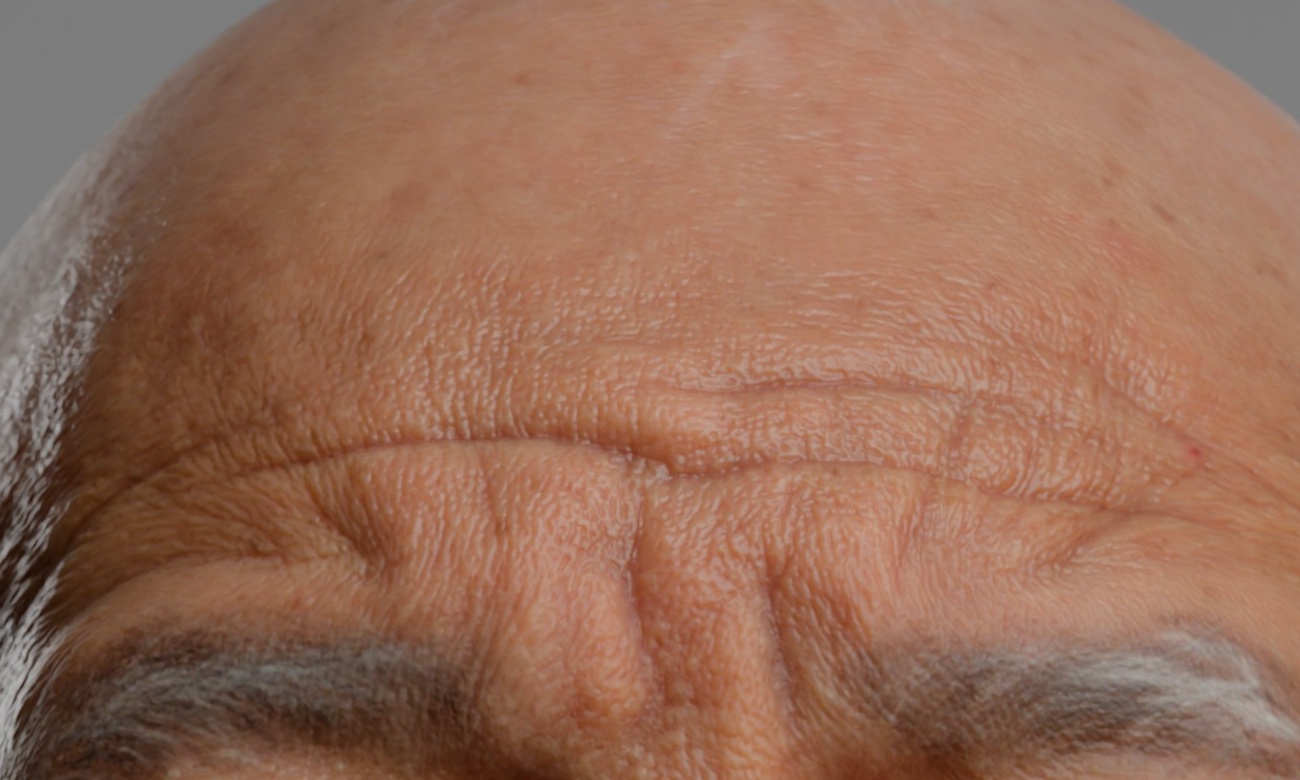} &
\includegraphics[width=3.35cm]{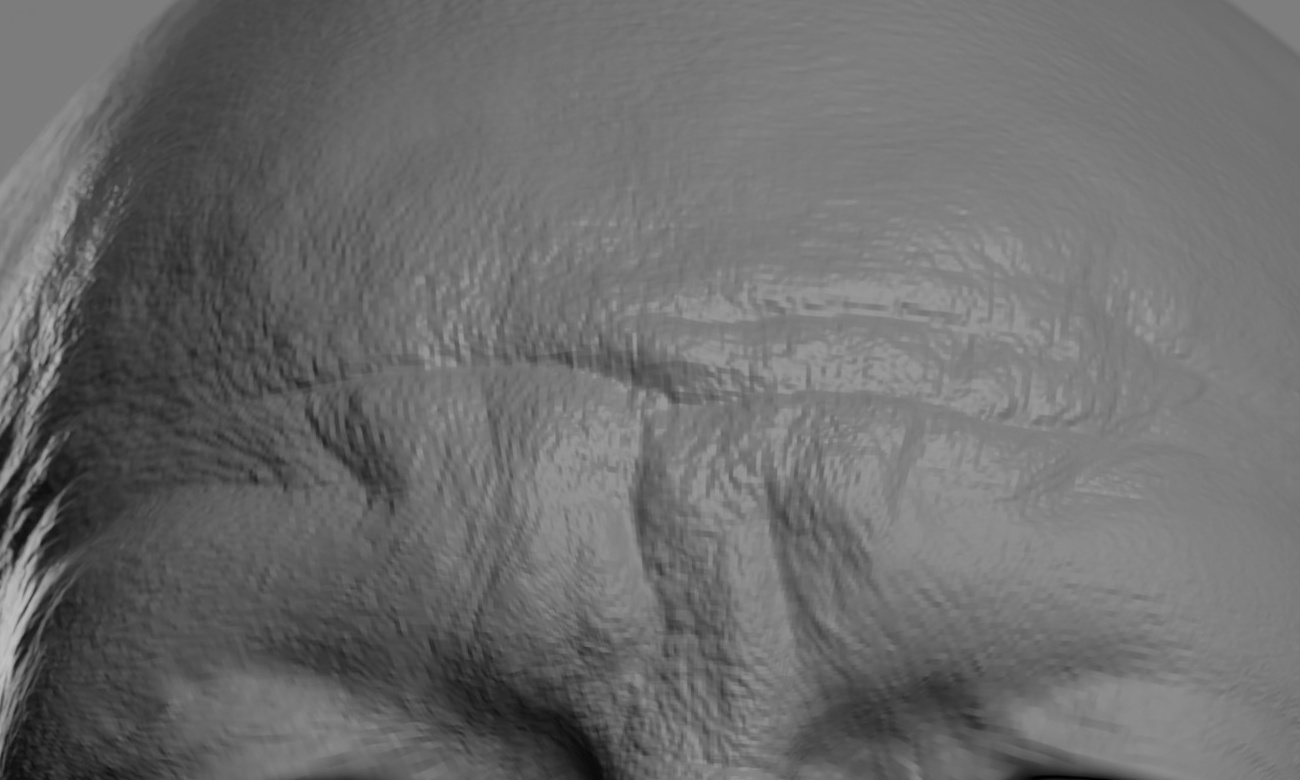} &
\includegraphics[width=3.35cm]{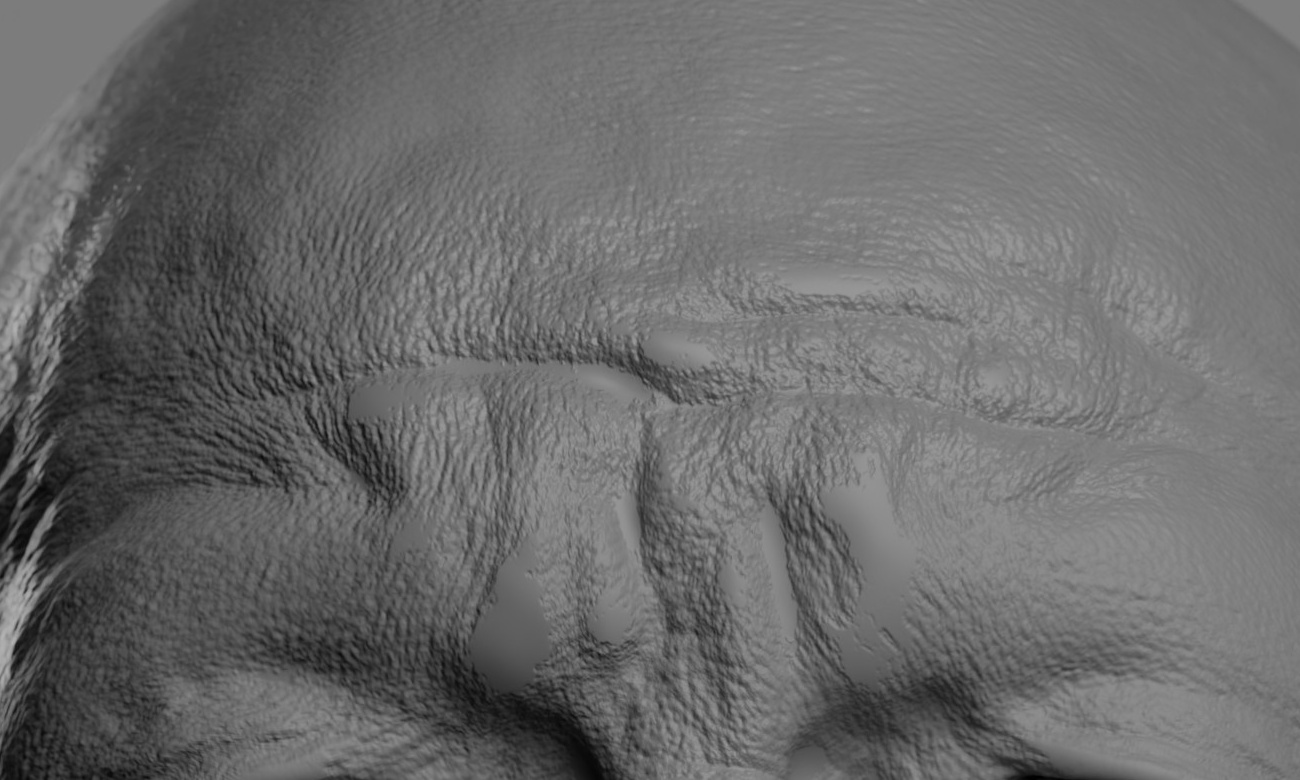} \\
\end{tabular}}
\caption{Domain-specific 4K super-resolution for a held-out map. Left to right:
1K/4K albedo and 1K/4K displacement, each with a matching forehead crop.
A light in-domain fine-tune recovers pore- and wrinkle-scale detail absent
from the $\RESLR{\times}\RESLR$ inputs. Rendered in Blender Cycles.}
\Description{For one subject: full renders (top) and zoomed crops (bottom) of
1K albedo, 4K albedo, 1K displacement, and 4K displacement.}
\label{fig:teaser}
\end{teaserfigure}

\maketitle
\renewcommand{\shortauthors}{Akhoundi et al.}

\section{Introduction}
\label{sec:introduction}

In modern digital-human workflows, physically based facial assets are represented by spatially registered UV maps, including an \emph{albedo} map for diffuse surface color and a \emph{displacement} map for mesoscopic geometry such as pores, fine expression lines, and micro-wrinkles.
Close-up rendering requires these maps at 4K resolution, yet generative pipelines often produce only 1K outputs that must be upscaled before production use.

Applying an off-the-shelf image super-resolution (SR) model to these maps is a poor fit.
Albedo UV layouts contain seams, charts, and near-uniform regions whose statistics differ substantially from natural photographs.
More fundamentally, displacement is a single-channel scalar height field dominated by high-frequency skin micro-structure, so photographic SR models can hallucinate visually plausible but geometrically incorrect relief.

We address this mismatch through lightweight in-domain adaptation rather than a new SR architecture.
For each modality, we adapt a frozen SDXL latent-diffusion backbone~\cite{podell2023sdxl,rombach2022ldm} as an image-to-image upscaler using LoRA modules~\cite{hu2021lora} on the UNet and a Tile ControlNet~\cite{zhang2023controlnet} conditioned on the low-resolution map.
This preserves the input structure while restoring domain-specific pore- and wrinkle-scale detail, with particular benefit for displacement, where generic photographic priors are most mismatched.

Our contributions are:
\begin{itemize}
\item A domain-tuned diffusion upscaler for facial UV assets that handles both RGB albedo and scalar displacement using lightweight adaptation of a frozen backbone.
\item A modality-specific displacement formulation that preserves the native geometric range in the latent path while providing a $[0,1]$ image condition to the Tile ControlNet.
\item A held-out evaluation showing sharper, more perceptually faithful 4K facial asset maps than bilinear and off-the-shelf GAN and diffusion super-resolvers.
\end{itemize}

\section{Related Work}
\label{sec:related}

\paragraph{Image super-resolution and diffusion priors.}
Single-image super-resolution has moved from CNN- and GAN-based restorers to diffusion priors: Real-ESRGAN~\cite{wang2021realesrgan} uses adversarial training with synthetic degradations, while SR3~\cite{saharia2023sr3}, StableSR~\cite{wang2024stablesr}, and DiffBIR~\cite{lin2024diffbir} synthesize detail from generative priors.
Their natural-image assumptions, however, are mismatched to our setting: facial UV asset maps contain seams, charts, and non-photographic statistics, and displacement is a scalar height field rather than an RGB appearance map.
We keep the pretrained-prior advantage of diffusion SR while adapting it to this production asset domain.

\paragraph{Controllable and parameter-efficient diffusion.}
Latent diffusion~\cite{rombach2022ldm} and SDXL~\cite{podell2023sdxl} provide strong high-resolution image priors, ControlNet~\cite{zhang2023controlnet} adds spatial conditioning for image-to-image refinement, and LoRA~\cite{hu2021lora} enables lightweight backbone adaptation.
We combine these in a targeted way: a frozen SDXL prior, Tile ControlNet to preserve the low-resolution UV structure, and per-modality LoRA adapters specializing the model for albedo and displacement.

\paragraph{UV texture and material-map generation.}
Recent work applies 2D generative priors to UV-space texture and material synthesis~\cite{richardson2023texture,yu2023pointuv,zeng2024paint3d,youwang2024paintit,zhang2024dreammat}, and DECA~\cite{feng2021deca} stresses the role of UV-space facial appearance and mesoscopic detail.
Most closely, PBR-SR~\cite{chen2025pbrsr} super-resolves PBR texture maps through a zero-shot differentiable-rendering optimization loop; we instead target facial albedo and displacement with a lightweight in-domain upscaler that needs no per-asset optimization at inference.

\section{Method}
\label{sec:method}

\begin{figure}[t]
\centering
\input{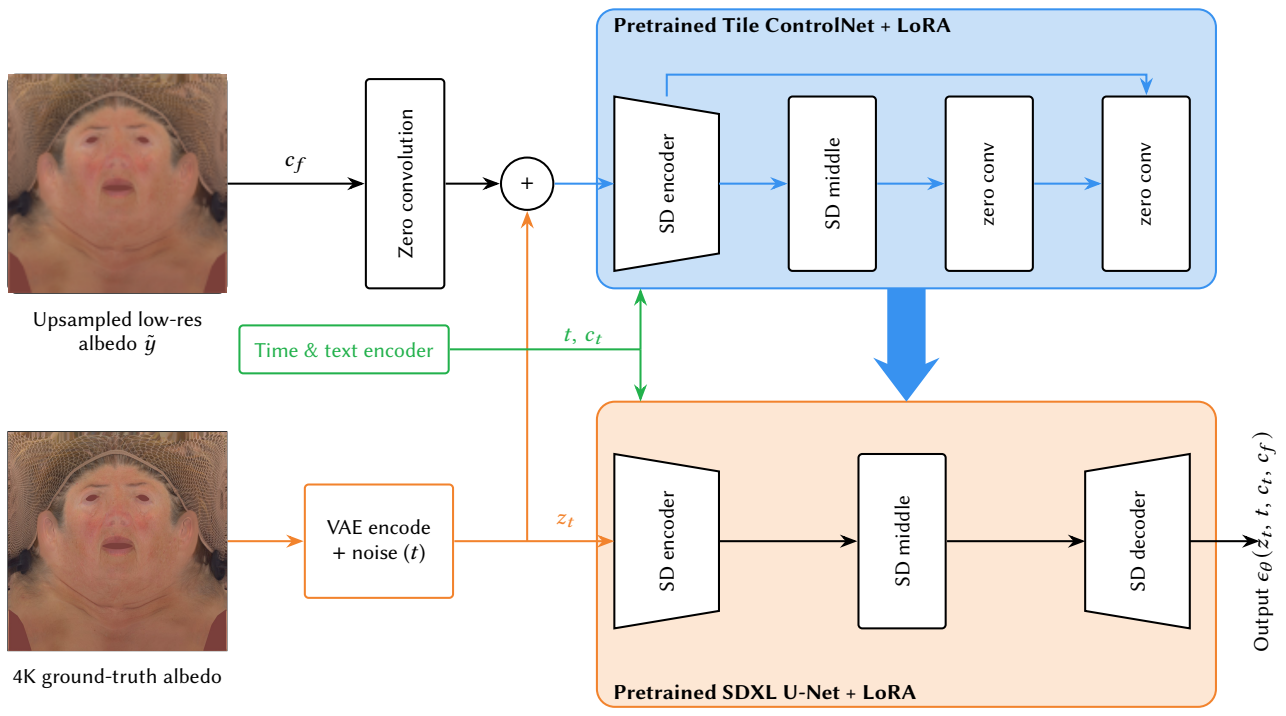}
\caption{Domain-tuned upscaler at \emph{training}. The 4K map is VAE-encoded
to $z_0$ and noised to $z_t$; a frozen SDXL UNet with LoRA predicts
$\epsilon_\theta(z_t,t,c_t,c_f)$, guided by a LoRA-adapted Tile ControlNet on
the bilinearly upsampled input $c_f$. Training uses aligned $128^2$ latent
tiles ($1024^2$ pixels), one model per modality. At \emph{inference}
(Sec.~\ref{sec:method}), the upsampled input initializes image-to-image
denoising; decoded albedo is chroma-matched to the input.}
\Description{Block diagram of the single-modality SDXL upscaler with Tile
ControlNet and LoRA adapters.}
\label{fig:pipeline}
\end{figure}

Our goal is to super-resolve a facial UV map from
$\RESLR{\times}\RESLR$ to $\RESHR{\times}\RESHR$ ($\SRFAC\times$)
for both albedo and displacement; we train one model per modality.
Given a low-resolution map $y$, we bilinearly upsample it to the target
resolution, $\tilde{y}=U(y)$, and refine it with a domain-tuned
image-to-image diffusion model rather than synthesizing from scratch.

\subsection{Image-to-image diffusion upscaler}
\label{sec:method-pipeline}
We use a pretrained SDXL~\cite{podell2023sdxl} latent-diffusion
backbone~\cite{rombach2022ldm} as a frozen generative prior, with its
base UNet weights, VAE, and text encoders frozen and adaptation performed
only through low-rank (LoRA) modules~\cite{hu2021lora}.
At \emph{training} time, the ground-truth high-resolution map $x$ is
VAE-encoded to $z_0$; we sample a timestep $t$, add noise to obtain $z_t$,
and train the LoRA-adapted UNet to predict the added noise, conditioned on
the aligned upsampled input.
At \emph{inference} time, we instead encode $\tilde{y}$, add noise to a
start step determined by the image-to-image strength, and denoise from
there, so the model refines $\tilde{y}$ rather than generating from noise.
A Tile ControlNet~\cite{zhang2023controlnet}, initialized from pretrained
Tile weights with frozen base weights and trained rank-$16$ LoRA adapters,
supplies spatial conditioning from $\tilde{y}$.
This keeps the output tied to the input UV layout while the diffusion prior
restores in-domain detail.

\subsection{Handling albedo and displacement}
\label{sec:method-modality}
Albedo is a three-channel RGB map in $[0,1]$, encoded using the standard
VAE preprocessing that maps it to $[-1,1]$.
Displacement is a single-channel scalar height field: a fixed global scale
maps it to $[-1,1]$, after which it is replicated across three channels for
the image VAE.
We avoid per-image photographic normalization so that the latent input
preserves the field's scalar range and ordering.
Because the pretrained Tile ControlNet expects a $[0,1]$ image condition,
we separately remap the displacement field to $[0,1]$ for conditioning.
Displacement therefore uses one representation for latent denoising and
another for spatial conditioning.
After decoding, the three predicted channels are averaged into a scalar
map and remapped to the original range.

\subsection{Training and inference details}
\label{sec:method-training}
We train for $5{,}000$ steps with an $L_1$ noise-prediction loss under
Min-SNR weighting ($\gamma{=}5$), using AdamW, an effective batch size of
$8$, and a cosine learning-rate schedule ($300$ warmup steps, peak
$5{\times}10^{-4}$, floor $10^{-5}$). The UNet uses rank-$64$ LoRA
adapters, while the Tile ControlNet uses rank-$16$ adapters at conditioning
scale $0.5$.
Each $\RESHR^2$ target is paired with a $\RESLR^2$ input obtained by
area-downsampling the 4K map by $\SRFAC$.
To make 4K training tractable, we cache each map's VAE latent posterior and
sample a $128^2$ latent tile per step, corresponding to a $1024^2$ pixel
region, together with the aligned crop from the upsampled input, so the VAE
is not invoked during training.
At inference, we refine $\tilde{y}$ with $50$ DPM-Solver++ steps at
strength $0.5$ and no classifier-free guidance.
Tiled VAE encoding and decoding process the full $\RESHR^2$ map, while the
UNet denoises the full latent in one pass, avoiding sliding-window seams.

\subsection{Albedo chroma preservation}
\label{sec:method-color}
For production albedo, we add a deterministic post-decode step to preserve
captured surface color.
We convert the decoded albedo and upsampled low-resolution input to CIELAB,
retain the decoded luminance ($L$), replace its opponent-color channels
($a,b$) with those of the input, and convert back to RGB.
Because chromatic variation is dominated by low frequencies, this anchors
surface color to the captured reference while retaining synthesized
luminance detail.
This step applies only to albedo.

\section{Results}
\label{sec:results}

\begin{figure}[t]
\centering
\includegraphics[width=0.74\columnwidth]{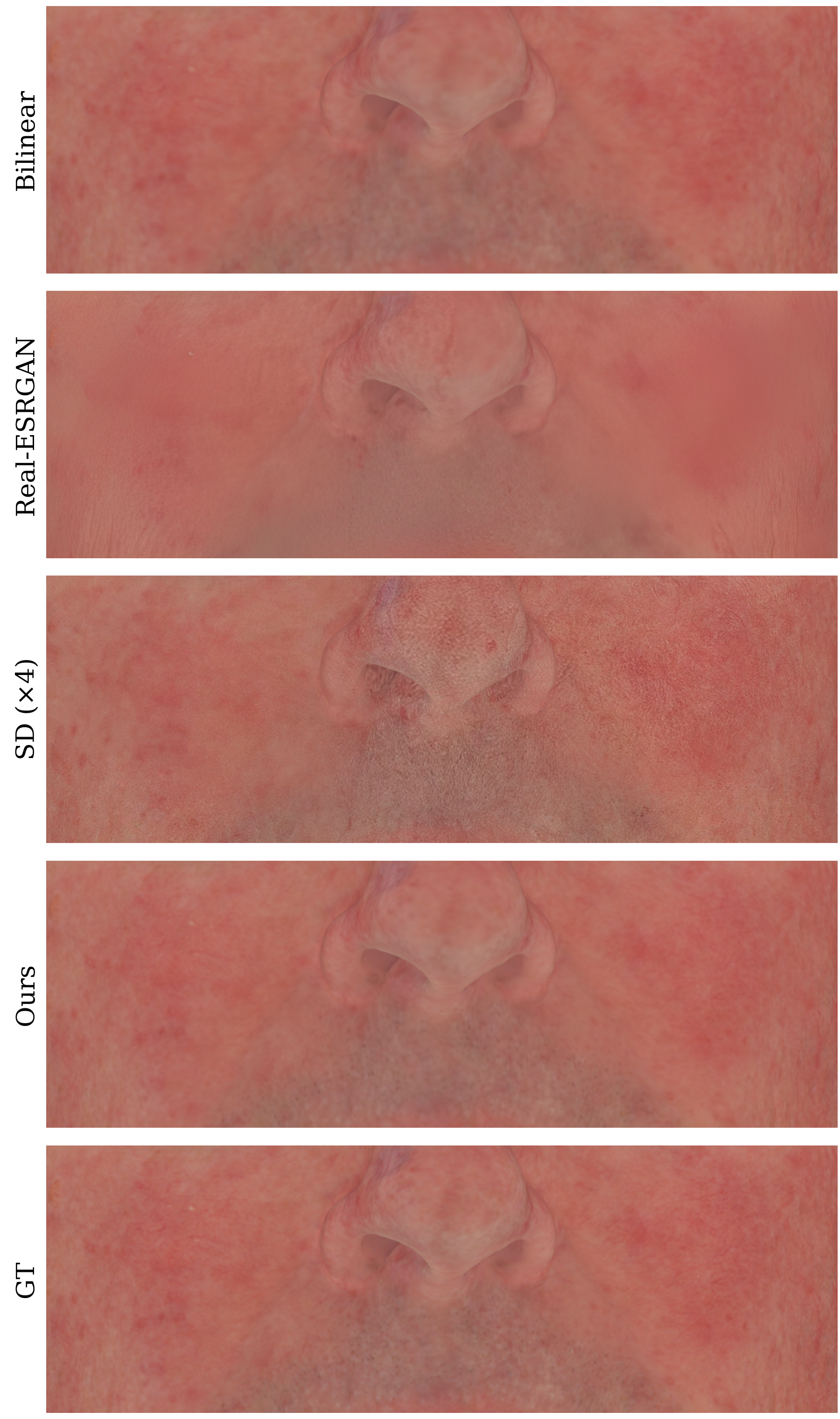}
\caption{\textbf{Albedo super-resolution ($\SRFAC\times$).}
Nose--mouth renders of each 4K albedo on the same head, with captured GT.
Bilinear and Real-ESRGAN flatten skin detail; SD ($\times4$) adds grainy,
off-domain texture. Ours most closely matches the reference.}
\Description{Five stacked nose--mouth renders comparing bilinear,
Real-ESRGAN, Stable Diffusion $\times4$, ours, and captured ground truth.}
\label{fig:detail}
\end{figure}

\begin{figure}[t]
\centering
\includegraphics[width=0.74\columnwidth]{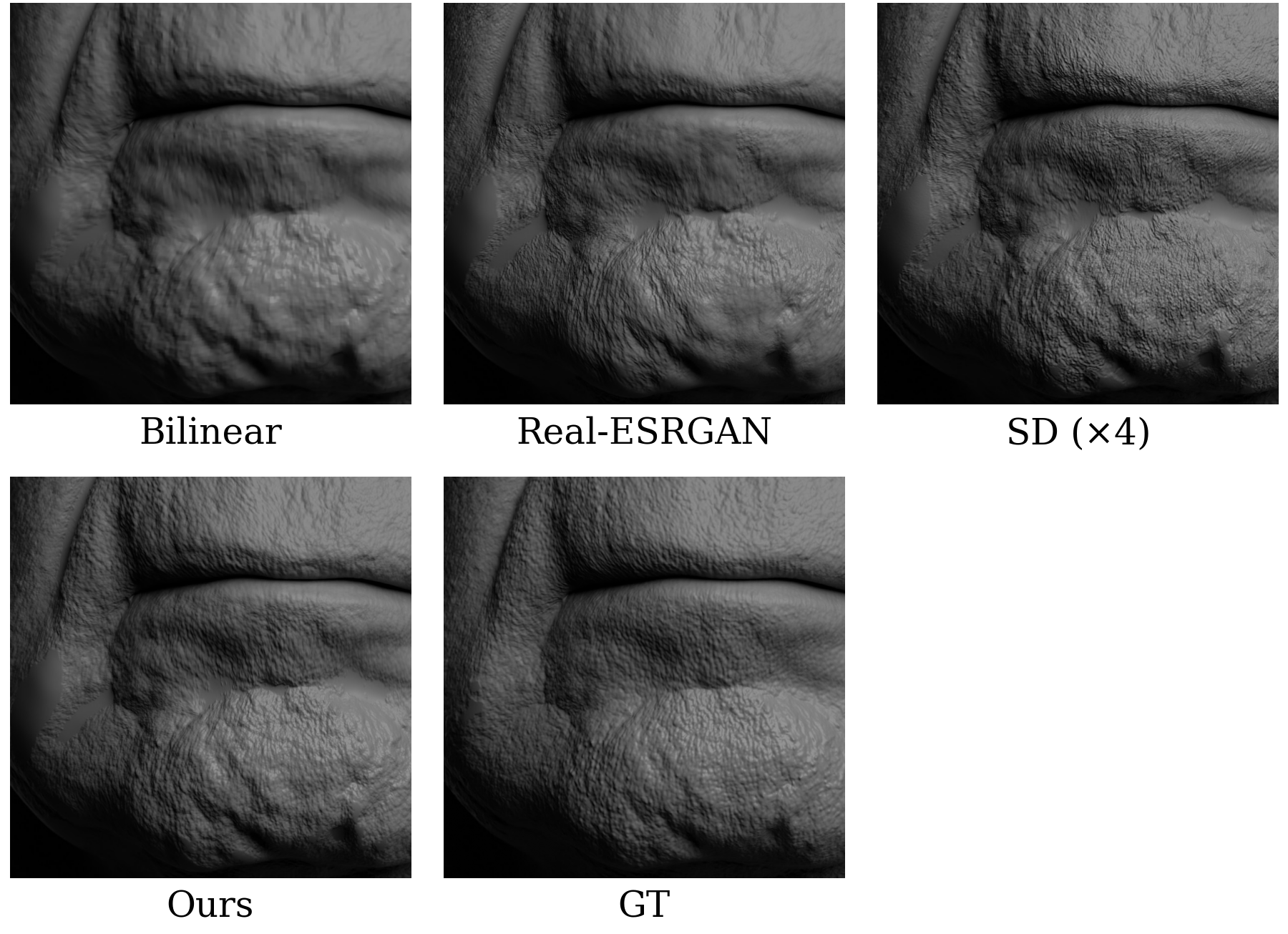}
\caption{\textbf{Displacement super-resolution ($\SRFAC\times$).}
Grey-clay mouth/chin renders of upscaled height fields, with captured GT.
Bilinear is blurry, Real-ESRGAN suppresses relief, and SD ($\times4$) adds
noise-like structure; ours is sharper and closer to GT
(Table~\ref{tab:main}). Blender Cycles, 64 spp, ACES.}
\Description{Five grey-clay mouth--chin renders comparing bilinear,
Real-ESRGAN, Stable Diffusion $\times4$, ours, and captured ground truth.}
\label{fig:detail_disp}
\end{figure}

\paragraph{Evaluation protocol.}
We evaluate on \NTEST{} held-out maps disjoint from training, super-resolving
each modality from $\RESLR^2$ to $\RESHR^2$ against 4K ground-truth maps from
a $28$-camera $12$MP rig. For albedo, we report PSNR, SSIM, and LPIPS. For
displacement, we also report a \emph{detail-band} score.
After subtracting a Gaussian-smoothed version of each map
($\sigma{=}2$\,px at $\RESHR$), we measure the predicted-to-ground-truth
detail-energy ratio ($1.0$ = correct magnitude) and Pearson correlation with
the ground-truth detail band ($1.0$ = perfect structural agreement).

\paragraph{Baselines.}
We compare each modality against bilinear upsampling, which initializes our
pipeline; Real-ESRGAN~\cite{wang2021realesrgan}; and the pretrained Stable
Diffusion $\times4$ upscaler~\cite{rombach2022ldm}. Both learned baselines are
used off-the-shelf, isolating the benefit of domain tuning. Displacement is
mapped to $[0,1]$ and replicated across three channels. All albedo methods
use the same CIELAB chroma-preservation step (Sec.~\ref{sec:method-color}).

\paragraph{Quantitative results.}
Table~\ref{tab:main} confirms that off-the-shelf super-resolvers are poorly
matched to this asset domain. Real-ESRGAN underperforms bilinear on standard
fidelity metrics for both modalities, suppressing albedo detail and distorting
fine displacement relief. SD ($\times4$) restores more high-frequency content,
but its grainy, off-domain texture produces the weakest albedo scores
(LPIPS $0.49$, SSIM $0.77$). Ours reduces perceptual error and more closely
matches the captured appearance (Fig.~\ref{fig:detail}).

PSNR and SSIM provide limited discrimination because their low-frequency bias
rewards smooth interpolation. Ours remains comparable to bilinear on these
metrics (albedo $35.7/0.94$; displacement $35.9/0.91$ vs.\
$36.2/0.93$), while reducing albedo LPIPS from $0.21$ to $0.08$.
For displacement, bilinear preserves alignment (correlation $0.61$) but
recovers only $46\%$ of ground-truth detail energy. Off-the-shelf models
overshoot the energy magnitude ($1.15$--$1.26$) with lower correlation
($0.33$--$0.36$). Ours recovers nearly the correct energy ($97\%$) while
preserving bilinear-level alignment ($0.61$). Since the 1K input does not
determine skin micro-structure, these results demonstrate structural
consistency rather than pixel-exact pore reconstruction.

\begin{table}[t]
\centering
\caption{$\SRFAC\times$ upscaling ($\RESLR{\to}\RESHR$) vs.\
4K GT ($\uparrow$/$\downarrow$ better). \emph{Detail}: recovered high-frequency
energy vs.\ GT ($\rightarrow\!1$ ideal). \emph{Corr}: agreement with the GT
detail band. Off-the-shelf methods overshoot and misplace displacement relief;
ours recovers near-correct energy with bilinear-level alignment.}
\label{tab:main}
\resizebox{\columnwidth}{!}{%
\begin{tabular}{lccccccc}
\toprule
 & \multicolumn{3}{c}{Albedo} & \multicolumn{4}{c}{Displacement} \\
\cmidrule(lr){2-4}\cmidrule(lr){5-8}
Method & PSNR\,$\uparrow$ & SSIM\,$\uparrow$ & LPIPS\,$\downarrow$
       & PSNR\,$\uparrow$ & SSIM\,$\uparrow$ & Detail\,($\rightarrow\!1$) & Corr\,$\uparrow$ \\
\midrule
Bilinear                                    & 35.69 & 0.94 & 0.21 & 36.22 & 0.93 & 0.46 & \textbf{0.61} \\
Real-ESRGAN~\cite{wang2021realesrgan}       & 29.69 & 0.86 & 0.25 & 31.12 & 0.81 & 1.26 & 0.33 \\
SD ($\times4$)~\cite{rombach2022ldm}        & 31.99 & 0.77 & 0.49 & 32.57 & 0.81 & 1.15 & 0.36 \\
Ours (fine-tuned)                           & 35.72 & 0.94 & 0.08 & 35.91 & 0.91 & \textbf{0.97} & \textbf{0.61} \\
\bottomrule
\end{tabular}%
}
\end{table}

\section{Discussion and Conclusion}
\label{sec:discussion}

We presented a domain-specific latent-diffusion upscaler that super-resolves
facial albedo and displacement maps from $\RESLR{\times}\RESLR$ to
$\RESHR{\times}\RESHR$ by adapting a frozen SDXL backbone with lightweight LoRA
modules and a Tile ControlNet conditioned on the low-resolution input.
Faithful $\SRFAC\times$ upscaling needs no custom SR architecture: in-domain
adaptation with modality-specific handling produces sharper, more faithful 4K
asset maps than off-the-shelf photograph-trained super-resolvers, particularly
for scalar displacement, and outputs render-ready maps in the asset UV layout
without per-asset optimization. The contribution is this domain
recipe---image-to-image initialization, Tile ControlNet, and scalar
displacement handling---not a new super-resolution architecture.

\paragraph{Limitations.}
Our evaluation uses held-out maps from a single capture pipeline at a fixed
$\SRFAC\times$ factor; robustness across capture conditions, scan qualities,
and other scale factors remains untested, and a different factor would require
retraining. We train separate models for albedo and displacement and do not
treat specular, roughness, or normal maps, nor do we enforce cross-map
consistency beyond the shared UV layout. Because a 1K input does not uniquely
determine skin micro-structure, recovered displacement is structurally
consistent rather than pixel-exact. We target production upscaling of 1K
generative or archived maps; cinematic VFX close-ups may still need artist
review.

\bibliographystyle{ACM-Reference-Format}
\bibliography{upscale}

\end{document}